\documentclass{article}
\usepackage[numbers,sort&compress]{natbib}

\usepackage[final]{nips_2018}

\usepackage[utf8]{inputenc} % allow utf-8 input
\usepackage[T1]{fontenc}    % use 8-bit T1 fonts
\usepackage{hyperref}       % hyperlinks
\usepackage{url}            % simple URL typesetting
\usepackage{booktabs}       % professional-quality tables
\usepackage{amsfonts}       % blackboard math symbols
\usepackage{nicefrac}       % compact symbols for 1/2, etc.
\usepackage{microtype}      % microtypography

\usepackage{xcolor}
\usepackage{amsmath}
\usepackage{booktabs}
\usepackage[final]{graphicx}
\usepackage{subcaption}

\title{Spatiotemporal Data Fusion for Precipitation Nowcasting}

\author{
  Vladimir Ivashkin\\
  Yandex, Moscow, Russia\\
  \texttt{vlivashkin@yandex-team.ru} \\
  \And
  Vadim Lebedev \\
  Yandex, Moscow, Russia \\
  \texttt{cygnus@yandex-team.ru} \\
}

\begin{document}

\maketitle

\begin{abstract}
Precipitation nowcasting using neural networks and ground-based radars has become one of the key components of modern weather prediction services, but it is limited to the regions covered by ground-based radars. Truly global precipitation nowcasting requires fusion of radar and satellite observations. We propose the data fusion pipeline based on computer vision techniques, including novel inpainting algorithm with soft masking.
%Image blending is a well-known task and can be solved using classical CV techniques and neural network approaches. In particular cases it is important to have both best quality and localness of blending. 
%To achieve changes only where mask allowed it, we refer to image inpainting methods. Existing inpainting methods using binary masks that cover corrupted areas. However, the damaged areas can be quite extensive and the neural network does not get enough information to restore the image. In this case, transparent masks can help: soft masks allow the transfer of some information about the corrupted area. We show qualitative and quantitative comparisons with other methods to confirm our approach.
\end{abstract}

\section{Introduction}

Precipitation nowcasting is a problem of precipitation prediction on a very short term (up to two hours). In this time range, extrapolation techniques working with detailed observational data can compete with and even win over meteorological models. The sources of data for this problem include ground-based radars and geostationary satellites. Radars provide more detailed and accurate observations, so using radars is more desirable when they are available. Nowcasting based on radar data was investigated in detail in literature~\cite{xingjian2015convolutional, shi2017deep}.

However, radar coverage is not complete across the globe. There are huge areas where only the satellite observations available, so any global solution for precipitation nowcasting have to incorporate satellite data and through the following steps

\begin{enumerate}
    \item Detect precipitation on satellite observations
    \item Perform fusion of radar and satellite data
    \item Train model and perform prediction
\end{enumerate}

The detection is described in~\cite{meyer2016comparison, tao2017precipitation} and approaches proposed for prediction step include extrapolation with optical flow~\cite{bowler2004development} and convolutional LSTM networks~\cite{xingjian2015convolutional}. In this paper, we focus on the fusion step of this pipeline that was not explored before. 

The fusion of radar and satellite observations is a challenging task, as two data domains differ both spatially (each radar detects precipitation in the range of 200 km around its location) and temporally (radars and satellites provide observations in different moments of time, with different cadence). We address temporal fusion by simple interpolation algorithm using optical flow, and spacial fusion (image blending) by alpha-blending and special modification of inpainting algorithm with a soft mask to correct artifacts.

In the simplest case of spatial fusion, the satellite data is replaced by radar data when the latter is available. As both radar and satellite have errors, this procedure produces sharp circular seams. In case of early fusion (fusion before prediction), the fixed in space seams will confuse the nowcasting model, which is supposed to estimate and predict the movement of precipitation fields. In case of late fusion, the seams will be left in the output map and confuse the users.

% nowcasting (contraction for now and forecasting) is a tool which shows people precipitation probability from now up to 2 hours in future. This is very useful tool for people because the rain does not become for them an unpleasant surprise. There are two source types about current rain field: weather radar data and weather satellite data. Radars detect rain by wave reflectivity and satellites use Earth pictures in different frequencies to detect rain clouds. On one hand, as far as radars directly detect a rain, radar data is more precise. On the other hand, radars have very low coverage in many countries.

%The optimal approach for precipitation nowcasting in regions without full radar coverage requires fusion of radar and satellite data, that have to address different spatial and temporal coverage of these sources. The fusion can be done before or after running of nowcasting algorithm, which is not covered in this work. Below, we address spacial and temporal fusion separately.

%The solution is to use radars where they exist and satellites where they not. Now the problem is how to blend satellite and radar data smooth: they have a different nature and rain fronts does not coincide on the border between radar and satellite source. Straight glue gives bad results, so we need an algorithm which can make smooth border and at the same time be local, because we care about precision inside domains.

Inpainting networks allow setting mask explicitly. Starting from neural network with partial convolutions \cite{liu2018image}, we improve inpaint quality by developing a new mechanism of mask updates. 

In summary, we make the following contributions:
\begin{itemize}
  \item We develop a global nowcasting using radar and satellite data.
  \item We develop a soft mask inpainting which outperform binary mask inpainting.
  \item We show that it will better variant for certain types of tasks like large area recovery and image blending.
\end{itemize}

%------------------------------------------------------------------------
\section{Related Work}

%We briefly review the relevant works from the classical image blending approach to modern approaches. We also discuss the difference between our work and the others.

%\subsection{Image blending}

%In classical computer vision, the theme of image mixing was developed and was well developed. The simplest algorithm is alpha blending \cite{uyttendaele2001eliminating}. 

Image blending is a well-developed topic of classical computer vision. Smooth transition between composited images can be guided by the pyramid image representation~\cite{adelson1984pyramid}. Optimal seam methods~\cite{milgram1975computer, efros2001image} enhance the visual quality of the composited image by cutting at optimal places, and Poisson image editing~\cite{perez2003poisson} corrects different exposure or tint of the composed images. Recently, GANs were applied to the problem \cite{wu2017gp}, allowing to make non-local photorealistic changes.

While our task essentially is the one of image blending, several factors stop us from adopting these methods: in our case, seam position is predefined by the radar range, there is no need for exposure correction, and it is desirable to make only the local changes.

Image inpainting, the process of reconstructing lost or corrupted regions of the image, can also be used for blending: compose two images together, declare the region near the seam to be corrupted, then inpaint. Multiple approaches for image inpainting were developed in the pre-CNN era of computer vision (i.e., reconstruction of isophotes~\cite{bertalmio2000image} or Patch-Match algorithm~\cite{Barnes:2009:PAR}), but the modern approaches rely on neural networks, their ability to learn the shapes and textures of the objects in the training set and reproduce them to cover corrupted parts of the image. These approaches commonly use GANs with rectangular binary masks to cover the 'corrupted' parts of the image~\cite{iizuka2017globally, yu2018generative}. Alternatively, \cite{liu2018image} propose a CNN architecture (inspired by UNet~\cite{ronneberger2015u} model) with the partial convolution layers and combination of losses which outperform \cite{iizuka2017globally, yu2018generative}.

The main idea behind partial convolution from \cite{liu2018image} is the following.
Let $\mathbf{W}$ be the convolutional weights and $b$ the corresponding bias. $\mathbf{X}$ are the pixels values for the current convolution window and $\mathbf{M}$ is the corresponding binary mask. The partial convolution at every location is expressed as:
\begin{equation}
  x' = \begin{cases}
    \frac{1}{sum(\mathbf{M})} \mathbf{W}^T(\mathbf{X} \odot \mathbf{M})  + b, & \text{if $sum(\mathbf{M}) > 0$}\\
    0, & \text{otherwise}
  \end{cases}
\end{equation}

where $\odot$ is an element-wise multiplication.
After each layer with the partial convolution, the mask values of the next layer $m'$ are updated:

\begin{equation}
  m' = \begin{cases}
    1, & \text{if $sum(\mathbf{M}) > 0$}\\
    0, & \text{otherwise}
  \end{cases}
  \label{eq:inpaint_mask}
\end{equation}

%------------------------------------------------------------------------
\section{Method}

%In this section, we first presenting our soft-mask improvement. We then present how this solution can be applied to domain blending.

%In this section, we describe our approach to spatiotemporal fusion of precipitation data.

\subsection{Temporal fusion}

The first problem we encounter during mixing satellite and radar data is a different time resolution. While most radars provide images once every 10 minutes, the time resolution for the satellite imagery (Meteosat-8) is 15 minutes. 

Most of the temporal change in the data is associated with precipitation fields moving in the wind, so the optical flow based algorithm comes up as a natural solution to produce intermediate frames required to link radar and satellite data. We use a simplified dense optical-flow based frame rate conversion algorithm~\cite{baker2011database} to generate synthetic satellite data at 10-minute intervals.

%Next, we compose radar and satellite images, blur the seam with alpha-blending and apply inpainting network, described below, to clean up the remains of the seam. The key feature of our approach is to use inpainting method with soft instead of binary masking.

%We follow an intuition, that inpainting network trained on smooth rain areas will tend to unite rain fronts. %Actually we can't measure inpainting quality because of we don't have ground truth data here.

%Both radar and satellite data has 3 classes of intensity. We use only satellite data during training and preprocess it using \ref{eq:training}. Here is some result images (the brighter color the more precipitation) 

\subsection{Soft-mask inpainting}
%All approaches in \ref{subsec:inpainting} use a binary mask to determine an area to fix.

%Based on the idea of partial convolutions~\cite{liu2018image}, we propose partial onvolutional Layer with non-binary masks. The intensity of every pixel in a mask shows confidence level of a correspondent pixel in a source image.

We extend partial convolutions~\cite{liu2018image} by allowing non-binary values in the mask, interpreted as the confidence levels of the corresponding pixels in a source image. With this modification, the inpainting algorithm incorporates some information from the corrupted part of the image and successfully fills larger regions.

%This modification expands the set of inpainting tasks to be solved. Binary masks algorithms can't fill large dense holes because of lack of information. Soft-mask approach help to use part of the information from the corrupted area.

%Proposal improvement is to make mask Equation \ref{eq:inpaint_mask} non-binary:
In case of smooth masking, Equation \ref{eq:inpaint_mask} changes the following way:

\begin{equation}
m' = \begin{cases}
    1, & \text{if $\max(\mathbf{M}) = 1$}\\
    m, & \text{otherwise}
  \end{cases}
\end{equation}

where $m$ is a correspondent element of the mask from the previous layer.
%If the previous layer is larger than a current layer because of dilation/pooling, one can apply the same transformation to $M$ to get appropriate $m$ in matrix form. 

%Resulting soft-mask layer allows us to propagate some information from picture under the mask.

\subsection{Soft-mask preprocessing for blending}
\label{subsec:preprocessing}

Since our inpainting model uses information from the area of blending, we preprocess the data to allow the network using both sources. During inference, we use simple alpha-blending to compose two images $I_1$ and $I_2$ with smooth transition between them

\begin{equation}
  I_{\text{inference}} = \alpha \odot I_1 + (1 - \alpha) \odot I_2
  \label{eq:inference}
\end{equation}

where values of the map $\alpha$ define this transition, linearly changing from $0$ to $1$ in the blending area. Note that ground truth can't be obtained for this scheme, so we need another approach for training.

During training, we use the satellite data $I$ as the ground truth, and apply noise $N$ to the simulated blending area to generate the training data $I_{\text{training}}$

\begin{equation}
  I_{\text{training}} = |1 - 2\alpha| \odot I + (1 - |1 - 2\alpha|) \odot N
  \label{eq:training}
\end{equation}

In this case, $\alpha$ is equal to zero in the middle of blending area, equal to one outside the blending area, and changes linearly between these two points.

%The results of the trained model are shown in Figure~\ref{fig:blendi}.

%As far as now network uses information from the blending area, we should preprocess data to allow the network to use both sources. For inference, we use simple alpha blending \cite{uyttendaele2001eliminating}, which can be defined in terms of margin.

%Assume that $\alpha(M)$ is a smooth monotonous symmetric function of margin M which takes $0$ in $M = -R$ and $1$ in $M = R$. R is a radius of blending. We have $I_1$ and $I_2$ images and want to make smooth transition 
%\begin{equation}
%  I_{\text{inference}}(p) = \alpha(M)I_1(p) + \alpha(-M)I_2(p)
%  \label{eq:inference}
%\end{equation}
%where $p = (x, y)$ point of image.

%For training we use one picture $I$ and noise image $N$ of the same shape. Every pixel of $N$ if randomly generated from uniform distribution.
%\begin{equation}
%  I_{\text{training}}(p) = 2(\alpha(|M|) - 0.5)I(p) + 2(\alpha(-|M|) - 0.5)N(p)
%  \label{eq:training}
%\end{equation}

%-------------------------------------------------------------------------
\section{Experiments}

\subsection{Comparison with binary mask inpainting}

%In this section, we want to measure how soft-mask approaches can outperform binary masks approaches.
In this section, we perform experiments on a different dataset to demonstrate the validity of our approach with soft masking.

We use CelebA-HQ \cite{liu2015deep, karras2017progressive} dataset for training and testing. For CelebA-HQ, we randomly partition into 27K images for training and 3K images for testing. All images scaled and cropped to 256x256 pixels.
For the training process, we add noise to the images in these datasets using Equation~\ref{eq:training} and try to predict source images. We use quick-draw~\cite{ha2017neural} as a shape of the mask.

We compare binary inpainting network and two soft-mask networks: one with a constant mask filled with value $0.5$ (semi-transparent mask) and another with mask $M = \alpha$, where $\alpha$ is taken directly from Equation \ref{eq:training}. We compare these networks using PSNR and SSIM metrics and present the results in Table~\ref{table:psnr_ssim_celeba}, demonstrating improvement of both variants of soft masking compared to binary masking.  

\begin{table}[h]
    \centering
    \caption{PSNR and SSIM metrics for inpainting networks trained on Celeba-HQ}
    \begin{tabular}{lcc}
     \toprule
     Inpainting       & PSNR & SSIM   \\
     \midrule
     Binary           & 32.6 & 0.9765 \\
     Semi-transparent & 35.3 & 0.9798 \\
     $\alpha$         & 35.9 & 0.9822 \\
     \bottomrule
    \end{tabular}
    
    \label{table:psnr_ssim_celeba}
\end{table}

The best performance is achieved in the case of $M = \alpha$, but in context of the task this is data leak.
% that can not be realized outside of experiments on synthetic data, because we don't have access to the value of $\alpha$ in reality.
However, the network with semi-transparent mask have almost the same metrics, demonstrating that the main improvement is caused by utilization of information from the masked region of the original image, and not from the leak of parameters of data synthesis through $\alpha$. Thus, we can use soft-mask inpainting with any reasonable $\alpha$ instead of exact values and still have an improvement in quality compared to binary masking.

\subsection{Radars and satellite data fusion}
\begin{figure}[h]
    \centering
    \begin{subfigure}{.19\textwidth}
      \centering
      \includegraphics[width=.9\linewidth]{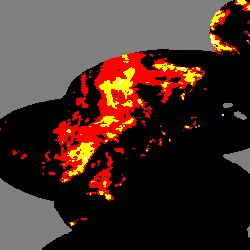}
    %   \caption{radars}
    \end{subfigure}
    \begin{subfigure}{.19\textwidth}
      \centering
      \includegraphics[width=.9\linewidth]{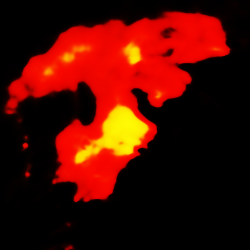}
    %   \caption{satellite}
    \end{subfigure}
    \begin{subfigure}{.19\textwidth}
      \centering
      \includegraphics[width=.9\linewidth]{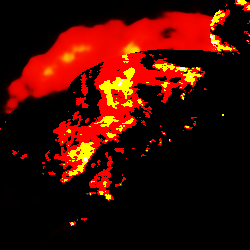}
    %   \caption{alpha blending}
    \end{subfigure}
    \begin{subfigure}{.19\textwidth}
      \centering
      \includegraphics[width=.9\linewidth]{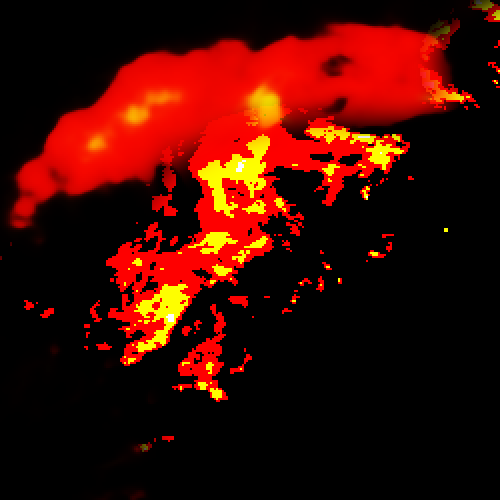}
    %   \caption{inpainting}
    \end{subfigure}
    \begin{subfigure}{.19\textwidth}
      \centering
      \includegraphics[width=.9\linewidth]{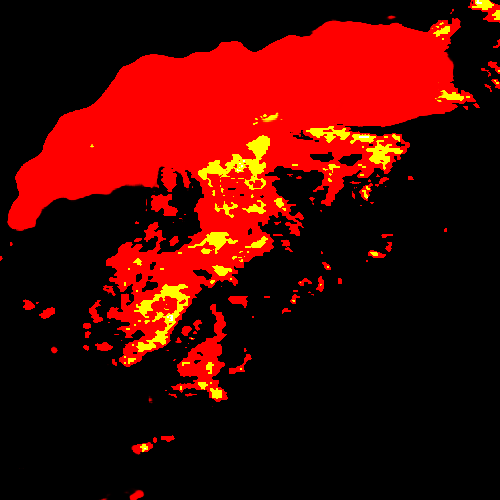}
    %   \caption{straight glue}
    \end{subfigure}\\
    
    \begin{subfigure}{.19\textwidth}
      \centering
      \includegraphics[width=.9\linewidth]{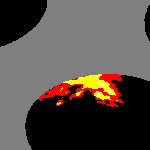}
    %   \caption{radars}
    \end{subfigure}
    \begin{subfigure}{.19\textwidth}
      \centering
      \includegraphics[width=.9\linewidth]{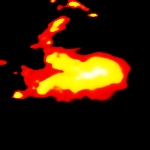}
    %   \caption{satellite}
    \end{subfigure}
    \begin{subfigure}{.19\textwidth}
      \centering
      \includegraphics[width=.9\linewidth]{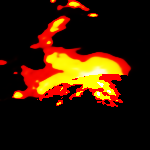}
    %   \caption{alpha blending}
    \end{subfigure}
    \begin{subfigure}{.19\textwidth}
      \centering
      \includegraphics[width=.9\linewidth]{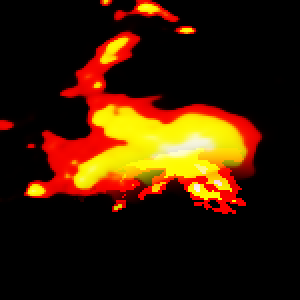}
    %   \caption{inpainting}
    \end{subfigure}
    \begin{subfigure}{.19\textwidth}
      \centering
      \includegraphics[width=.9\linewidth]{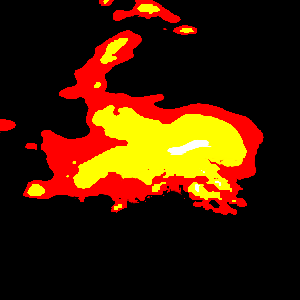}
    %   \caption{straight glue}
    \end{subfigure}\\
    
    \begin{subfigure}{.19\textwidth}
      \centering
      \includegraphics[width=.9\linewidth]{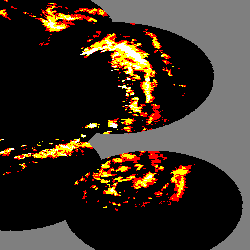}
    %   \caption{radars}
    \end{subfigure}
    \begin{subfigure}{.19\textwidth}
      \centering
      \includegraphics[width=.9\linewidth]{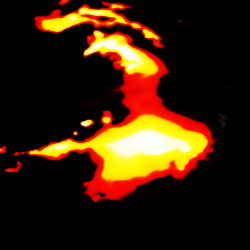}
    %   \caption{satellite}
    \end{subfigure}
    \begin{subfigure}{.19\textwidth}
      \centering
      \includegraphics[width=.9\linewidth]{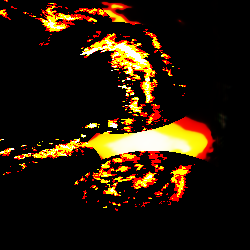}
    %   \caption{alpha blending}
    \end{subfigure}
    \begin{subfigure}{.19\textwidth}
      \centering
      \includegraphics[width=.9\linewidth]{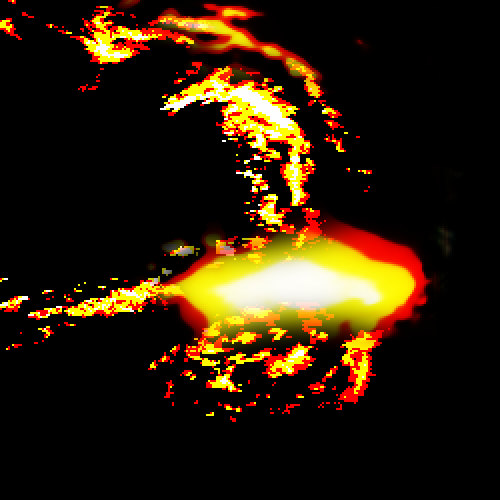}
    %   \caption{inpainting}
    \end{subfigure}
    \begin{subfigure}{.19\textwidth}
      \centering
      \includegraphics[width=.9\linewidth]{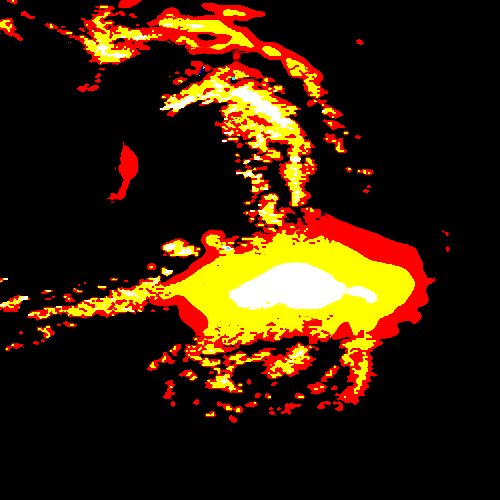}
    %   \caption{straight glue}
    \end{subfigure}\\
    
    \begin{subfigure}{.19\textwidth}
      \centering
      \includegraphics[width=.9\linewidth]{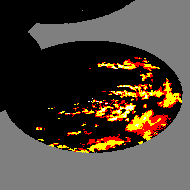}
    %   \caption{radars}
    \end{subfigure}
    \begin{subfigure}{.19\textwidth}
      \centering
      \includegraphics[width=.9\linewidth]{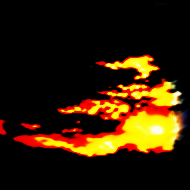}
    %   \caption{satellite}
    \end{subfigure}
    \begin{subfigure}{.19\textwidth}
      \centering
      \includegraphics[width=.9\linewidth]{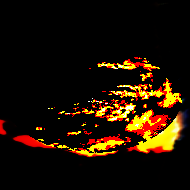}
    %   \caption{alpha blending}
    \end{subfigure}
    \begin{subfigure}{.19\textwidth}
      \centering
      \includegraphics[width=.9\linewidth]{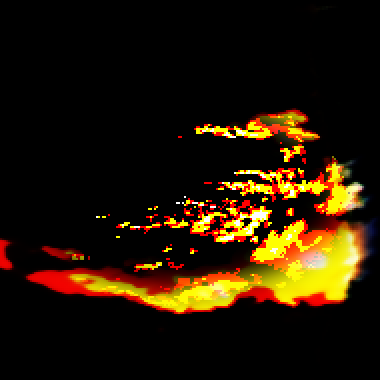}
    %   \caption{inpainting}
    \end{subfigure}
    \begin{subfigure}{.19\textwidth}
      \centering
      \includegraphics[width=.9\linewidth]{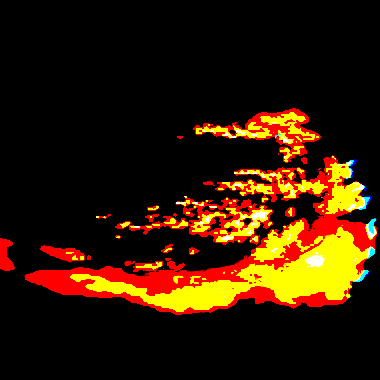}
    %   \caption{straight glue}
    \end{subfigure}\\
    
    \begin{subfigure}{.19\textwidth}
      \centering
      \includegraphics[width=.9\linewidth]{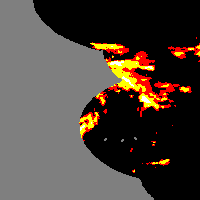}
      \caption{radars}
    \end{subfigure}
    \begin{subfigure}{.19\textwidth}
      \centering
      \includegraphics[width=.9\linewidth]{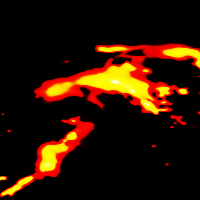}
      \caption{satellite}
    \end{subfigure}
    \begin{subfigure}{.19\textwidth}
      \centering
      \includegraphics[width=.9\linewidth]{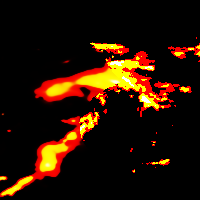}
      \caption{no blending}
    \end{subfigure}
    \begin{subfigure}{.19\textwidth}
      \centering
      \includegraphics[width=.9\linewidth]{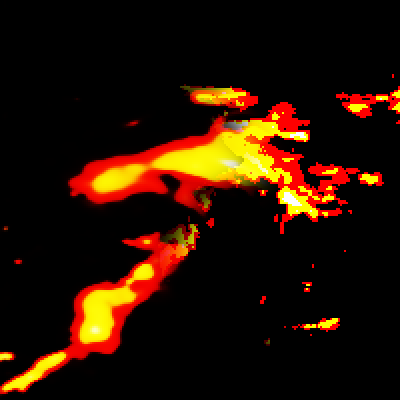}
      \caption{alpha blending}
    \end{subfigure}
    \begin{subfigure}{.19\textwidth}
      \centering
      \includegraphics[width=.9\linewidth]{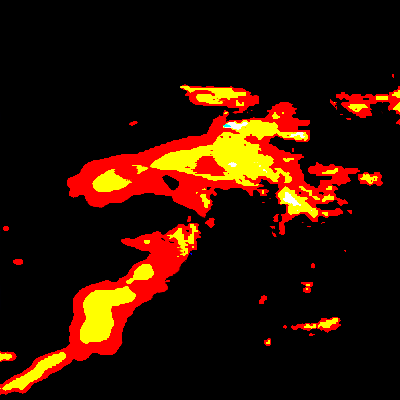}
      \caption{inpainting}
    \end{subfigure}\\
    
    \caption{Results of data fusion in a spatial domain. The first panel demonstrates radar data (notice the partial coverage), the second panel shows satellite observations, third - results of the composition without any form of blending, with a sharp edge between two types of data. In the fourth column, the edge is smoothed by alpha blending, and in the last column, it is further cleaned by our inpainting procedure}
    \label{fig:blendi}
\end{figure}

Figure \ref{fig:blendi} shows results of the same model on our target task, satellite and radar data fusion. It can be seen that our approach successfully hides the border between two types of data.

%-------------------------------------------------------------------------
\section{Conclusion}

We presented a pipeline for spatiotemporal fusion of precipitation data. Our pipeline uses simple techniques, such as optical flow based interpolation and alpha-blending, as well as a novel approach of image inpainting with soft masking. We have demonstrated an advantage of soft masking in the case then useful information can be recovered from the corrupted region. The proposed pipeline allows expanding precipitation nowcasting from the list of selected locations covered with radars (USA, Hong Kong, Europe), to the rest of the world. 

%We presented a simple but novel technique that can be useful for restoration images with large semi-damaged areas. Also, we shown image blending can be solved as an inpainting task.

{\small
\bibliographystyle{unsrt}
\bibliography{egbib2}
}

\end{document}